\pdfoutput=1
\documentclass[11pt]{article}

\usepackage{acl}

\usepackage{times}
\usepackage{latexsym}

\usepackage[T1]{fontenc}
\usepackage[utf8]{inputenc}

\usepackage{microtype}

\usepackage{inconsolata}

\usepackage{graphicx} % Required for inserting images
\usepackage{caption}
\usepackage{subcaption}
\graphicspath{ {./graphs/} }
\usepackage[utf8]{inputenc}

\usepackage{amsmath,amsfonts,mathtools}
\usepackage{xcolor}
\usepackage{hyperref}
\usepackage{algorithmic}
\usepackage{algorithm}
\usepackage{ulem}
\usepackage{multirow}
\usepackage{booktabs}

\newcommand{\be}{\begin{equation}}
\newcommand{\ee}{\end{equation}}

\def\[{\LLeft[}
\def\]{\right]}
\def\({\LLeft(}
\def\){\right)}
\def\<{\LLangle}
\def\>{\rangle}

\usepackage[colorinlistoftodos,prependcaption,textsize=small]{todonotes}

\DeclareCaptionType{Example}

\title{Few-shot Policy (de)composition in Conversational Question Answering}

 \author{Kyle Erwin \and Guy Axelrod \and Maria  Chang \and  Achille Fokoue \\ \and {\bf Maxwell Crouse } 
  \and {\bf Soham Dan } \and {\bf Tian Gao } \and  {\bf Rosario Uceda-Sosa} \\
  \and {\bf Ndivhuwo Makondo } \and {\bf Naweed Khan } \and {\bf Alexander Gray} }  
\begin{document}
\maketitle

\begin{abstract}
The task of policy compliance detection (PCD) is to determine if a scenario is in compliance with respect to a set of written policies. In a conversational setting, the results of PCD can indicate if clarifying questions must be asked to determine compliance status.
%The task of policy compliance detection (PCD) is to determining the compliance of scenarios to written policies or other legal frameworks, with potential options to ask followup questions to gather further information. 
Existing approaches usually claim to have reasoning capabilities that are latent or require a large amount of annotated data. In this work, we propose logical decomposition for policy compliance (LDPC): a neuro-symbolic framework to detect policy compliance using large language models (LLMs) in a few-shot setting. By selecting only a few exemplars alongside recently developed prompting techniques, we demonstrate that our approach soundly reasons about policy compliance conversations by extracting sub-questions to be answered, assigning truth values from contextual information, and explicitly producing a set of logic statements from the given policies. %The formulation of explicit logic graphs can in turn help answer policy compliance detection (PCD) questions with increased transparency, explainability, and accuracy. We demonstrate the superiority of our proposed approach over multiple existing baselines on the popular PCD benchmark, ShARC.
The formulation of explicit logic graphs can in turn help answer PCD-related questions with increased transparency and explainability. We apply this approach to the popular PCD and conversational machine reading benchmark, ShARC, and show competitive performance with no task-specific fine-tuning. We also leverage the inherently interpretable architecture of LDPC to understand where errors occur, revealing ambiguities in the ShARC dataset and highlighting the challenges involved with reasoning for conversational question answering.
\end{abstract}

\section{Introduction}
% todo: Introdiction to be written after all other sections have been written

%This is a planning document for work on policy compliance, to improve \citep{saeidi2021cross} on the Question Answering for Policy Compliance (QA4PC) dataset.
Policy compliance detection (PCD) is the task of automatically determining if a scenario complies with or violates a policy. Automated methods for PCD can be used across a variety of domains, e.g. detecting regulatory violations, determining benefits eligibility, or enforcing social media guidelines. In most real-world applications, scenarios are typically underspecified, making it difficult to determine compliance without obtaining additional information. In a conversational setting, additional information can be obtained by generating follow-up questions in a task referred to as conversational machine reading \cite{saeidi-etal-2018-interpretation}. One way to approach the PCD task is to decompose policies into questions and combine those questions into logical expression trees that can be used to arrive at the final answer \cite{saeidi2021cross}. %Recently, \cite{kotonya2022policy} show that expression trees may be created without requiring labeled expression trees for training.   %. This type of PCD paired with conversational natural language has been referred to as conversational machine reading , since the machine reading comprehension system must interpret the scenario and determine what additional information to seek. 
%The task of policy compliance detection (PCD) is to determining the compliance of scenarios to written policies or other legal frameworks. An instantiation of PCD is the case in which a scenario is presented alongside a policy, and we need to determine if a specific situation adheres to the rules defined in the text of the policy. In some cases, one can also ask followup questions to users to further gather the necessary information to resolve ambiguity. In this case, the task is also called conversational machine reading (CMR) \cite{saeidi2018interpretation}, with both document interpretation and dialog QA understanding component. 

Large language models (LLMs) %, particularly decoder-only models such as GPT and LLaMA,
are good candidates for conversational PCD, given their impressive natural language generation abilities that resemble deep understanding and emergent reasoning. However, LLMs have been shown to have issues with factuality \cite{ji2023survey} and reasoning \cite{valmeekam2022large}. %This is problematic for most tasks, but especially for PCD tasks that may be applied to high stakes decision support systems. 
Many approaches seek to improve factuality and reasoning capabilities by having LLMs ``show their work'' \cite{wei2022chain,yao2024tree}, though even these elicited ``thoughts'' may not accurately reflect the factors that influence LLMs outputs \cite{turpin2024language}. To mitigate the risks of incorrect or unfaithful explanations, a variety of self-improving LLMs have been proposed \cite{wang2022self,pan2023automatically} as a way to decompose the opaque nature of LLM inference.

In this paper, we propose Logical Decomposition for Policy Compliance (LDPC), which uses LLM-based reasoning that can be applied to the PCD task. Specifically, our system decomposes a natural language policy into yes/no questions that can be composed into an explicit logical formula to be used for PCD and for asking follow up questions. Importantly, the PCD decision is deterministically inferred from the logical formula, as opposed to being generated in an opaque manner and tenuously connected to an explanation or chain of thought.

 \noindent We make the following contributions in this paper:
\begin{enumerate}
    \item We present Logical Decomposition for Policy Compliance (LDPC): an inherently interpretable approach to conversational PCD that combines neural question decomposition and logic formulation with symbolic reasoning.
    \item Our policy decomposition and logical formulation modules receive no task-specific fine-tuning, yet achieve competitive results on the ShARC dataset.
    \item We propose a simple but powerful three-valued logic that is especially suitable for conversational PCD.
    \item We identify ambiguities in the ShARC dataset and challenges that arise from reasoning about policies to answer questions in a conversational setting
\end{enumerate}

\section{Related Work}\label{sec:related-work}
This paper builds upon prior work in two general areas: conversational machine reading (Section \ref{sec:related-work:cmr}) and reasoning with large langauge models via prompt-based instruction (Section \ref{sec:related-work:prompting}).

\subsection{Conversational Machine Reading} \label{sec:related-work:cmr}
% \td{shorten and add more relevant works}
Conversational machine reading is a type of question answering task that requires the joint interpretation of a scenario and some background information to answer a question, where the appropriate response to the question may not be an answer at all, but rather, a follow-up question \cite{saeidi-etal-2018-interpretation}. It was introduced as a more challenging and realistic alternative to question answering tasks where answers were spans of text that could be extracted from a paragraph, as in the highly influential Stanford Question Answering Dataset \cite{rajpurkar2016squad}. We leverage two influential datasets in this domain: ShARC \cite{saeidi-etal-2018-interpretation} and QA4PC \cite{saeidi2021cross}.

\subsubsection{ShARC}\label{sec:related-work:sharc} The ShARC dataset \cite{saeidi-etal-2018-interpretation} is a dialogue-based question-answering dataset constructed from 948 distinct text snippets. These snippets are taken from government websites and describe a policy. Each snippet features an input question and a corresponding ``dialog tree.'' These trees branch based on yes/no answers to follow-up questions at each step. The dataset comprises all individual branching paths, referred to as ``utterances'', from these trees. There are 6,058 utterances in total. Additionally, 6,637 scenarios provide extra information about the user, enabling the skipping of certain questions. Scenarios modify dialog trees, resulting in 37,087 distinct utterances when combined with negative sampled scenarios. After removing unreachable portions of dialog trees, the final dataset consists of 32,436 utterances. For each utterance, a model must predict one of the following classes: irrelevant, yes, no, follow-up question. If the model decides that a follow up question is the most appropriate response, it must also generate that follow up question. The dataset is divided into train, development, and test sets, with sizes of 21,890, 2,270, and 8,276, respectively. Evaluation metrics include Micro and Macro Accuracy and BLEU. Micro and Macro Accuracy assess the model's ability to classify each utterance, while BLEU evaluates the quality of generated questions.

Many techniques have been introduced to compete for top performance on ShARC. For example, \cite{zhong-zettlemoyer-2019-e3} train a model to extract rules from policy text and determine which rules are entailed by the conversation history. Using self-attention layers for the input and extracted rules, a decision module is able to learn how to rephrase rules into follow up questions and when to ask those questions. In contrast, \cite{lawrence-etal-2019-attending} use bidirectional attention to make predictions on ShARC. The dialogue graph modeling (DGM) algorithm \citep{ouyang2020dialogue} improves on Discern \cite{gao2020discern} by using explicit discourse graph and implicit discourse graph models with graph convolution network (GCN). ET5 \citep{zhang2022et5} proposes a end-to-end framework with T5 to fully exploit the entailment reasoning in decision making steps by using a shared encoder (with a  multi-task architecture) and separate decoders for reasoning and answer generation. T-reasoner \citep{sun2022scenario} integrates a reasoning module to model condition relationships and verify consistent answers within user scenarios, identifying unsatisfied conditions. It breaks up the passage into individual conditions, which are then checked by the reasoning module. However, progress on ShARC must be interpreted judiciously, as \cite{vermaneural} find spurious clues and patterns that make trained models susceptible to artificial gains in performance.

 %\footnote{https://github.com/ozyyshr/DGM}
 %\footnote{https://github.com/Yottaxx/ET5}
% \footnote{https://github.com/haitian-sun/T-Reasoner} 

%in the ShARC \citep{saeidi-etal-2018-interpretation} dataset and proposes a fix to remove patterns that result a simple heuristic would go well. 
 %Finally, \cite{verma-etal-2020-neural}%\footnote{https://github.com/nikhilweee/neural-conv-qa}
%studied the spurious clues and patterns in the ShARC \citep{saeidi-etal-2018-interpretation} dataset and proposes a fix to remove patterns that result a simple heuristic would go well. 

\subsubsection{QA4PC}\label{sec:related-work:qa4pc}
Question answering for policy compliance (QA4PC) \cite{saeidi2021cross} augments 30\% of the ShARC dataset with questions to policies and expression trees. Each expression tree consists of a set of questions and logical operators that combine the answers to determine a final answer to a user question. By using the provided expression trees, a model explicitly checks each condition to obtain the answers, with improved accuracy and transfer capability (with new policies in testing). Logic Expression Tree \citep{kotonya2022policy} infers expression trees automatically from policy texts with a few supervised models. Authors introduce constrained decoding using a finite state automaton to ensure the generation of valid trees. Overall, these papers aim to improve the ability of language models to perform complex reasoning and decision-making tasks through training supervised by ShARC and QA4PC datasets. %is a dataset built upon ShARC. QA4PC addresses the task of policy compliance detection by decomposing a policy into an expression tree. %The objective of the dataset is 1) for a system to answer the given questions from the user scenario, and 2) and arrange the answers to those questions as an expression tree that can be evaluated to answer the user's question.

% \subsection{Learning with Logics}\label{sec:related-work:lwlg}
% % \td{add more related works, see Reasoning with LLMs survey paper -- Ndivhuwo will share in slack}
% %\td{maybe a few citations on logics w/o llm?}

% Utilizing explicit logic reasoning have been shown to improve the accuracy and consistency of language tasks [CITE].  In policy compliance, \cite{saeidi2021cross} augments 30\% of the ShARC dataset with questions to policies and expression trees. By using the provided express trees, a model explicitly checks each condition to obtain the answers, with improved accuracy and transfer capability (with new policies in testing). Logic Expression Tree \citep{kotonya2022policy} infers expression trees automatically from policy texts with a few supervised models. Authors introduce constrained decoding using a finite state automaton to ensure the generation of valid trees. Overall, these papers aim to improve the ability of language models to perform complex reasoning and decision-making tasks, which can be useful in various applications such as natural language processing, question answering, and dialogue systems.

\subsection{Reasoning with LLMs via Prompting}\label{sec:related-work:prompting}
%\td{general discussion of prompting and fine-tuning with citations -- Max, Ndivhuwo}

Many recent works have explored prompting strategies to address complex reasoning problems. These methods avoid finetuning altogether and instead use elaborate prompts to elicit from the LLM an explicitly written out, step-by-step solution to the problem. In the original chain-of-thought \cite{wei2022chain} work, prompting involved demonstrating a number of examples (i.e., few-shot prompting) that had questions paired with reasoning traces. 

The work of \cite{Kojima2022LargeLM} showed that prompting the LLM to produce a reasoning trace could be done without any few-shot examples whatsoever by simply initializing the prompt with the words ``Let's think step-by-step''. Soon after, more elaborate zero-shot prompting strategies like Plan-and-Solve prompting \cite{Wang2023PlanandSolvePI} and ReACT \cite{yao2022react} were introduced, where the LLM was prompted to solve the problem by first planning out a sequence of steps to follow, and then solving those steps in the order as described. 

Self-consistency \cite{wang2022self} is a decoding technique designed to leverage the notion that multiple alternative reasoning paths may lead to the correct answer. In the self-consistency framework, a diverse set of outputs are first sampled from the LLM. Then, the response returned to the user is the \textit{answer} most commonly produced by the LLM after stripping away the chain of reasoning produced in the response. This is equivalent to marginalizing over the alternative reasoning paths to select the most consistently produced answer. 

For dataset creation, the Orca \cite{mitra2023orca} framework explored tuning a smaller model with synthetic data produced by a model that had been prompted with a number of different problem-solving prompts. 
Self-taught Reasoner (STaR) \citep{zelikman2022star} proposes a technique to iteratively leverage a small number of example rationales to bootstrap reasoning capabilities of LLMs. It first includes a few example rationales in its prompts, and for wrong answers, tries to derive the answer and rationale again given the correct answers, and lastly fine-tunes on all the rationales that led to correct answers.  %\url{https://github.com/ezelikman/STaR}

% \td{general discussion on Logic learning and generations, especially with (self-taught or automatically generated) LLM -- Max, Ndivhuwo}

% \td{a few papers on LLM + Logics -- Max, Ndivhuwo}

% Reason and Act (ReAct) \citep{yao2022react} explore the use of LLMs to generate rasoning traces and task-specific actions in an interleaved manner. Human annotators provide thoughts and fixed actions (with API calls), and prompts the LLMs to follow. \url{https://github.com/ysymyth/ReAct}

\section{Problem Formulation}\label{sec:problem-formulation}

%\subsection{Problem Statement}
% \td{Fix this to be consistent with Algorithm notation}
Given a dataset $\mathcal{D}$ with size $D = |\mathcal{D}|$ with input $X = \{P, C\}$ and target output $O$, where $P = \{p_i\}_{i=1}^D$ is a set of policy documents, and $C = \{( q_i, s_i, h_i)\}_{i=1}^D$ is each context for the corresponding policy document $p_i$, containing user scenarios $s_i$, user question $q_i$, and dialogue history $h_i$. We wish to answer each input question  $q_i$ with either \textit{Yes}, \textit{No}, \textit{Irrelevant}, or follow up question $\texttt{questions}_i$. Hence, the  output of the model is $O = \{(\texttt{answer} _i, \texttt{questions} _i)\}_{i=1}^D$, containing the decision $\texttt{answer}_i$ on policy compliance and followup questions $\texttt{questions}_i$ when the decision $\texttt{answer}_i$ requires so. %Each $\texttt{questions}_i$ contains questions and their labels.
We will drop index $i$ to simplify notations when there is no ambiguity.

% [not needed?]
% In addition, let $G = \{g_i\}_{i=1}^D$ be the set of logic expression graphs for each data sample $i$. Additionally, assume we have access to a set of data with labeled logic graphs $\mathcal{D}_l^{\text{rat}} = \{(x_j,y_j, g_j)\}_{j=1}^L$ with size $L$.

%Given a ShARC item consisting of a natural language policy snippet $p$, an input question $q$ regarding policy compliance, a scenario $s$ and a dialogue history $h$, we wish to answer the input question with either Yes, No, Irrelevant or a follow up question.

% \begin{enumerate}
%     \item Logic graph generations: (self-taught or automatically generated) with LLM
% \item general discussion of prompting and fine-tuning
% % \item Uncertainty quantification: logical graphs
% \end{enumerate}

%\subsection{Potential Novelties.}
%% \paragraph{Dataset Cleaning.} remove data leakage

%\paragraph{LLMs.}
%only in-context training (no training) with intermediate steps
%(related works: STaR, ReACT)
%fine tuning with Pre-training LLM 

% \paragraph{Neural models.}
% argument mining
% semantic detection

%\paragraph{Learning Logic decomposition tree.}
%automatically learn the decomposition tree
%incorporation of logical graphs

%\paragraph{Few examples needed.} Few-shot learning setting
% \paragraph{Uncertainties.}
% uncertainties detection
% incorporation credal sets for uncertainties 

\section{Proposed Approach}\label{sec:proposed}

%This section details the proposed approach, which is inspired by the question answering for policy compliance (QA4PC) pipeline \citet{kotonya2022policy}, in that it decomposes a policy into a set of questions that can be arranged into logical formulae for reasoning. 
We propose a neuro-symbolic system that uses an LLM to decompose a piece of policy text into distinct conditions that are phrased as simple yes/no questions. The LLM then re-composes those conditions into symbolic logical formulae, which can be deterministically reasoned over using a simple three-valued logic and evaluated into an appropriate conversational response (illustrated with an example in Figure \ref{fig:overview}). 
Our pipeline has the following components that we describe below: (1) question-policy relevance, (2) policy decomposition, (3) logical formulation, (4) logical evaluation.

%1) A semantic similarity check on whether the question asked is relevant to the policy present. If relevant, 2) a process referred to as %question policy decomposition is used to generate questions about a given policy. 3) These questions are then organized into a logical formula, where answers to the questions serve as truth values. 4) The formula is then evaluated to produce a final answer.

\subsection{Question-Policy Relevance}\label{sec:proposed:semantic}
The first step in the proposed pipeline is to determine whether the user question is relevant to the policy. This is achieved by using the all-MiniLM-L12-v2 sentence transformer\footnote{https://huggingface.co/sentence-transformers/all-MiniLM-L12-v2}, which maps sentences and paragraphs to a 384-dimensional dense vector space. The question and policy are transformed into this vector space, and the cosine similarity between the sentence embeddings of the policy and question is calculated. If the similarity exceeds a threshold predetermined from a sensitivity analysis, the user question is considered relevant. Otherwise, the pipeline returns \textit{Irrelevant}.

\subsection{Policy Decomposition}\label{sec:proposed:pd}
\begin{figure}
    \centering
    \includegraphics[width=0.45\textwidth]{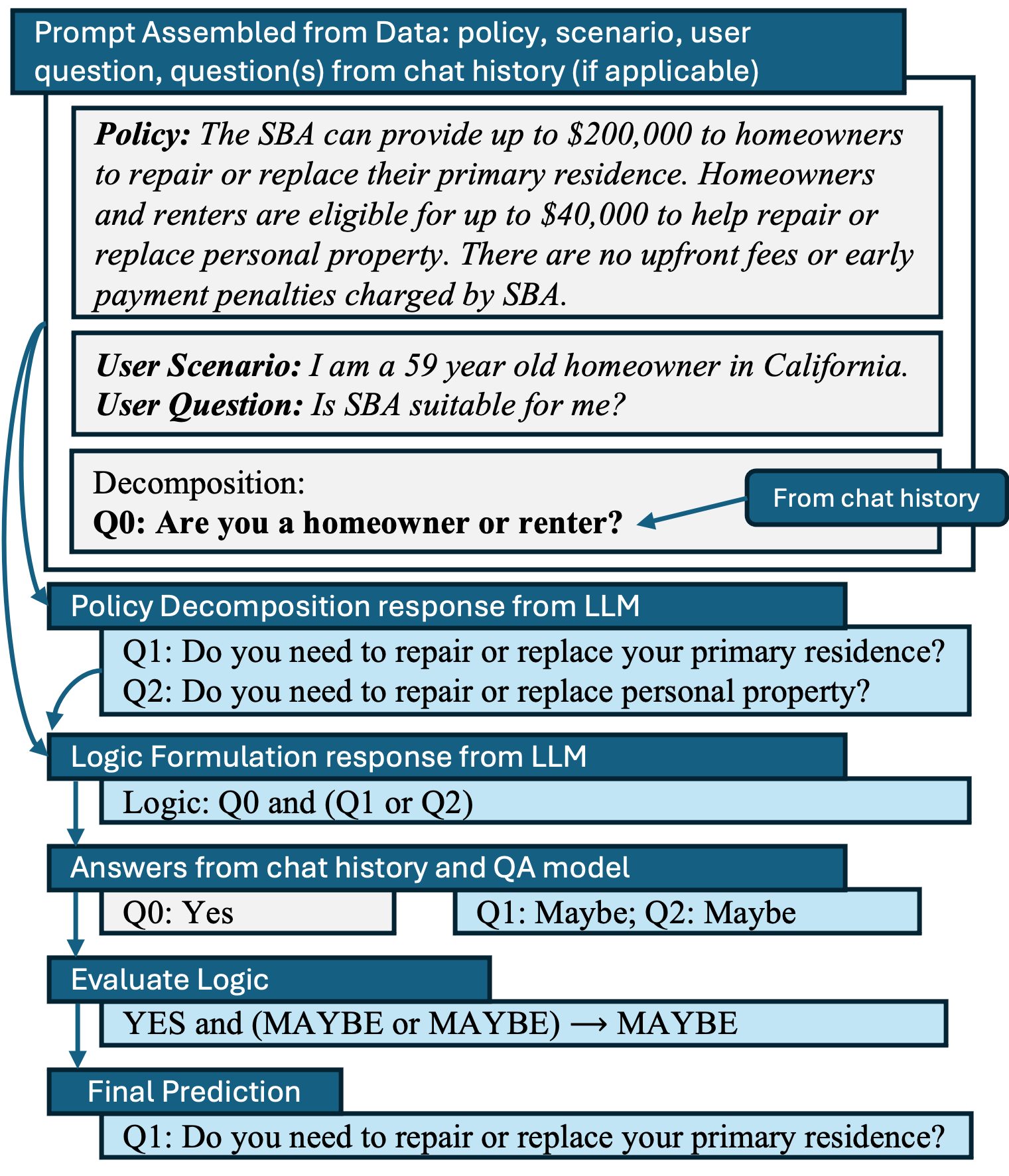}
    \caption{Policy decomposition and logical formulation given a policy, scenario, user question, and chat history (if applicable).}
    \label{fig:overview}
\end{figure}

In the policy decomposition step, an LLM, prompted with 20 in-context examples, breaks down the policy into basic yes or no questions that need to be answered in order to answer the user's question. The in-context examples are taken from the QA4PC dev set, since QA4PC includes decomposed questions from policies and the logical representations of those policies. The in-context examples are followed by the ShARC dataset item on which we want the model to perform inference.
%The questions are labeled as Q0, Q1, Q2, and so on. Existing yes or no-type questions from the chat history are included in the prompt to avoid the LLM from generating questions that have already been asked. 
The example in Figure \ref{fig:overview} shows how a prompt is constructed from the policy, scenario, user question. This particular example has one question from the chat history, so it is included as Q0. The model then produces the remaining questions Q1 and Q2, which represent potential follow up questions that may be asked to the user. %Example 1 provides an example prompt for the policy decomposition step. The prompt includes the policy, the user's question and a single question (Q0) from the chat history. The model's responds with additional questions, i.e. Q1, Q2, and Q3, to ask the user. These additional questions are shown in bold. The prompt shown in Example 1 is preceded by the instruction and in-context examples (see Appendix).

% \begin{figure}\label{example:pd}
% 	\centering
%     \fbox{
%         \begin{minipage}{7cm}
% \textit{Policy: Eligible applicants may obtain direct loans for up to a maximum indebtedness of \$300,000, and guaranteed loans for up to a maximum indebtedness of \$1,392,000 (amount adjusted annually for inflation).\\User: Does this loan meet my needs?\\Decomposition:}\\
% \textit{Q0: Do you need a direct loan?}\\
% \textbf{Q1: Do you need a guaranteed loan?}\\
% \textbf{Q2: Do you need to borrow less than \$300,000?}\\
% \textbf{Q3: Do you need to borrow less than \$1,392,000?}
%         \end{minipage}
%     }
%     \captionof{Example}{Policy decomposition given policy, user question and chat history. LLM output is in bold.}
% \end{figure}

\paragraph{Question Relevance}\label{sec:proposed:qr}
To ensure accurate and relevant responses during the policy decomposition stage, a filtering process is implemented to eliminate non-pertinent questions generated by the LLM. This process takes into account the user's query, chat history, and policy to prevent unnecessarily lengthy logical structures and varying logics for the same input. The LLM is asked to determine the relevance of each generated question, responding with a binary \textit{Yes} or \textit{No} answer. If the response is \textit{No}, the question is filtered out. This process is repeated for each generated question.

\subsection{Logic Formulation}\label{sec:proposed:lf}
At the logic formulation stage, the model is prompted with the input and output of the policy decomposition stage and creates a logical formula that can be used to answer the user's question. We again adopt a few-shot in-context learning strategy to generate the logical formulae via an LLM. The LLM is asked to ``combine the question variables into a python boolean expression'' (for full instructions see Appendix). The valid logic operators used are \texttt{and}, \texttt{or}, and \texttt{not}. Figure \ref{fig:overview} shows how this works on a single example. Note that the resulting logical formula includes question IDs from the policy decomposition stage (Q1 and Q2) as well as the question ID given by the conversation history (Q0).

% \begin{figure}
% 	\centering
%     \fbox{
%         \begin{minipage}{7cm}
% \textit{Policy: Eligible applicants may obtain direct loans for up to a maximum indebtedness of \$300,000, and guaranteed loans for up to a maximum indebtedness of \$1,392,000 (amount adjusted annually for inflation).\\User: Does this loan meet my needs?\\Decomposition:}\\
% \textit{Q0: Do you need a direct loan?}\\
% \textit{Q1: Do you need a guaranteed loan?}\\
% \textit{Q2: Do you need to borrow less than \$300,000?}\\
% \textit{Q3: Do you need to borrow less than \$1,392,000?}\\
% \textit{Logic:} \textbf{(Q0 and Q2) or (Q1 and Q3)}
%         \end{minipage}
%     }
%     \captionof{Example}{Logic formulation decomposition given policy, user question, chat history and generated questions. LLM output is in bold.}
% \end{figure}

\paragraph{Self-Consistency}

%Prompting strategies that elicit lines of reasoning from an LLM (e.g., chain-of-thought \cite{wei2022chain}, ReACT \cite{yao2022react}) have proven to be quite useful when applying LLMs to solve complex problems. When an LLM utilizes such a prompting strategy, it may output any of a number of reasoning paths that justify its solution. Self-consistency \cite{wang2022self} is a decoding technique designed to leverage the notion that multiple alternative reasoning paths may lead to the correct answer. 

%In the self-consistency framework, a diverse set of outputs are first sampled from the LLM. Then, the response returned to the user is the \textit{answer} most commonly produced by the LLM after stripping away the chain of reasoning produced in the response. This is equivalent to marginalizing over the alternative reasoning paths to select the most consistently produced answer. 

In this work, we adopt self-consistency \cite{wang2022self} during the generation of logical forms. In particular, rather than greedily decoding one logical form for a particular policy, our approach first samples $k$ alternative logical forms from the LLM. The logical forms are then grouped together into equivalence classes, i.e., all logical forms that are logically equivalent are put into the same partition. Last, the returned logical form is then selected to be one of the constituents of the largest partition of the sampled logical forms (our approach takes the logical form from the selected partition that is shortest). 

As an example, consider a situation where the LLM produced three logical forms: $\neg (A \wedge B)$, $(\neg A \vee \neg B)$, and $(A \wedge B)$. First, the sampled expressions would be grouped into buckets based on logical equivalence, leading to two sets $S_1 = \{ \neg (A \wedge B), (\neg A \vee \neg B) \}$ and $S_2 = \{ (A \wedge B) \}$. Then, the bucket selected would be the one with the largest size, in this case being $S_1$. From $S_1$, the shortest logical form (in terms of number of symbols) would be returned, thus resulting in $\neg (A \wedge B)$ being the returned expression.

\subsection{Logic Evaluation}

In our approach, we evaluate the generated propositional formula using a three-valued logic that introduces a \textit{Maybe} ($m$) value to indicate an unanswered question. The particular three-valued logic we use is commonly referred to as \textit{Kleene’s strong three-valued logic} in the non-classical logic literature\footnote{First introduced by Stephen Kleene in \cite{kleene-1938}. See \cite{Bergmann_2008} for an overview.}. 

\noindent Our set of truth values is $\mathcal{T} = \{F,m,T\}$, and we assume an ordering where $F < m < T$.\\
We consider the operations $\neg,\land,\lor$ on $\mathcal{T}$ where $\neg := \{(T,F),(F,T),(m,m)\}$ and for all $a,b \in \mathcal{T}$,
\begin{align*}
    a \land b := min(a,b)\\
    a \lor b := max(a,b)
\end{align*}

\noindent A conjunction operator evaluates to \textit{Maybe} if the operands are \textit{True} and \textit{Maybe} (or both \textit{Maybe}); the disjunction operator evaluates to \textit{Maybe} if the operands are \textit{False} and \textit{Maybe} (or both \textit{Maybe}), and the negation of \textit{Maybe} is \textit{Maybe}. Otherwise, all operands work as expected in binary logic.  The truth table for these operations is shown in Table \ref{tab:truth_table}.

This approach allows us to effectively handle uncertain or incomplete information, which is particularly useful in question and answering systems. If the generated logic evaluates to \textit{Maybe}, it signals that a follow up question is required. If the generated logic evaluates to \textit{True} or \textit{False}, it means that the answer to the user's question is \textit{Yes} or \textit{No}, respectively. We use the following question-answering process discussed below to determine the values for logical variables.

\subsection{Question Answering}

% \td{QUESTION FOR KYLE/SOHAM: did we stick to the approach below or did we use RoBERTa fine tuned ONLY on MNLI?
% Kyle is working on new results with model only trained on MNLI
% ...Note to Kyle: if we end up saying this part is truly independent of ShARC, you can rephrase the 2nd contribution in the intro to be more general}

We use a RoBERTa-large model fine-tuned on the MNLI corpus and further fine-tune on the ShARC NLI data. We fine-tuned for $5$ epochs, using a batch size of $32$, learning rate of $1e^{-5}$, and weight decay of $0.1$.
% \td{Kyle: I have added some text on how we use the fine tuned model}
We used the fine-tuned entailment model to determine if the user scenario can answer a generated question. To do this, we first convert the generated questions into statements. We then pass the user scenarios and the statements to the fine-tuned model, which determine whether the scenario entails the statement. In other words, the model determines whether the scenario provides sufficient information to answer the question. The model returns one of three possible values based on its evaluation: ``entails,'' ``contradiction,'' or ``neutral.'' We map these values to truth values for our logical evaluation. Specifically, ``entails'' corresponds to \textit{True}, ``contradiction'' corresponds to \textit{False}, and ``neutral'' corresponds to \textit{Maybe}. 

In Figure \ref{fig:overview}, this approach is used to determine the answers to Q1 and Q2. The answer to Q0 is provided in the chat history. With these truth values, the logical formula evaluates to \textit{Maybe}, which indicates that a follow-up question should be asked, leading to the final prediction, ``Do you need to repair or replace your primary residence?'' Note that both Q1 and Q2 are acceptable responses for this example.

\begin{table}[]
\begin{center}
\small
\begin{tabular}{ |c|c|c|c|c| } 
\toprule
\textbf{Q0} & \textbf{Q1} & \textbf{Q0 $\land$ Q1} & \textbf{Q0 $\lor$ Q1} & \textbf{$\neg$ Q0} \\ 
\midrule
\texttt{TRUE} & \texttt{TRUE} & \texttt{TRUE} & \texttt{TRUE} & \texttt{FALSE} \\
\midrule
\texttt{TRUE} & \texttt{FALSE} & \texttt{FALSE} & \texttt{TRUE} & \texttt{FALSE} \\
\midrule
\texttt{TRUE} & \texttt{MAYBE} & \texttt{MAYBE} & \texttt{TRUE} & \texttt{FALSE} \\
\midrule
\texttt{FALSE} & \texttt{FALSE} & \texttt{FALSE} & \texttt{FALSE} & \texttt{TRUE} \\
\midrule
\texttt{FALSE} & \texttt{MAYBE} & \texttt{FALSE} & \texttt{MAYBE} & \texttt{TRUE} \\
\midrule
\texttt{MAYBE} & \texttt{MAYBE} & \texttt{MAYBE} & \texttt{MAYBE} & \texttt{MAYBE} \\
\bottomrule
\end{tabular}
\end{center}
\caption{Truth table for operations in the three-val\-ued logic used in our approach.}
\label{tab:truth_table}
\end{table}

% Let $\Phi$ denote a countable set of propositional variables (in our case this is the set of all question labels that could occur in a generated logic formula). The language of the logic we are concerned with is $\mathcal{L} = \Phi \cup \{\textbf{and},\textbf{or},\textbf{not}\}$ and the set of well formed formulas $Frm(\mathcal{L})$ is given in BNF by
% $$\varphi ::= p \: | \: \varphi_1 \textbf{ and } \varphi_2 \: | \: \varphi_1 \textbf{ or }  \varphi_2 \: | \: \textbf{not } \varphi_1$$
% where $p \in \Phi$.\\
% A \textbf{truth-assignment/valuation} is a mapping $v: \Phi \to \mathcal{T}$. A valuation $v$ can be uniquely extended to the \textbf{evaluation} $\overline{v} : Frm(\mathcal{L}) \to \mathcal{L}$ which satisfies the following for all $\varphi, \psi \in Frm(\mathcal{L})$,
% \begin{itemize}
%     \item $\overline{v}(p) = v(p)$ for every $p \in \Phi$.
%     \item $\overline{v}(\varphi \textbf{ and } \psi) = \overline{v}(\varphi) \land \overline{v}(\psi)$
%     \item $\overline{v}(\varphi \textbf{ or } \psi) = \overline{v}(\varphi) \lor \overline{v}(\psi)$
%     \item $\overline{v}(\textbf{not }\varphi) = \neg \overline{v}(\varphi)$
% \end{itemize}
% Given a truth-assignment $v$ and formula $\varphi \in Frm(\mathcal{L})$, evaluating $\varphi$ refers to determining the value of $\overline{v}(\varphi)$.

% \td{Pushed code for OSS here \url{https://github.ibm.com/IBM-Research-AI/ControllableFoundationModels/tree/os-code}. It can be cleaned up a little more, but it is not urgent.}
\subsection{Overall Algorithm}

Algorithm \ref{alg:pipeline} outlines our approach's pipeline. The pipeline takes in $\mathcal{I}$, a tuple containing the policy ($p$), the user question ($q$), the user scenario ($s$), and the chat history ($h$). The pipeline consists of several functions that take in textual inputs, format them, and pass them to their respective language models to generate output. Specifically, $\mathcal{S}$ determines the semantic similarity between the user question and policies, while $\mathcal{D}$ represents the question decomposition process that takes in a LLM ($M_\mathcal{D}$), policy, user question, and chat history, and returns generated questions and their IDs as a dictionary. $\mathcal{Q}$ represents the question answering step, which takes in the user scenario and question, and $\mathcal{R}$ determines if a generated question is relevant to the policy and user question, and makes sense given the chat history. Finally, $\mathcal{L}$ represents the call to a LLM ($M_\mathcal{L}$) to generate the logic formula that best represents the policy.

Lines 6-13 assign answers to corresponding question IDs, with answers being taken from the chat history if the question is present, or determined using $\mathcal{Q}$. Lines 14-23 represent the question filtering process in cases where there are too many questions that may complicate the next step of logical formulation for a model. Lines 24-32 represent the logical formulation step in a self consistency loop. Finally, an answer is returned on line 33.
% \td{label -> id}
%=================================================

\begin{algorithm}[h]
    \caption{Logical Decomposition for Question Answering}
    \small
    \begin{algorithmic}[1]
        % \STATE \textbf{Require:} Models $\mathcal{D}, \mathcal{L}$; $M_q$
        \STATE \textbf{Input:} $\mathcal{I} = (p,q,s,h)$
        
        \IF{$\mathcal{S}(p, q) < 0.25$} 
            \RETURN \texttt{``Irrelevant"} 
        \ENDIF
        
        \STATE $\texttt{questions} \leftarrow  h \cup \mathcal{D}(M_\mathcal{D},p,q,h)$
        \STATE $\texttt{answers} \leftarrow  \{\}$
        \FOR{$(\texttt{ID}, \texttt{question}) \in \texttt{questions}$}
            \IF{$\texttt{ID} \in h$}
                \STATE $\texttt{answers[ID]} \leftarrow h[\texttt{ID}][``answer"]$
            \ELSE
                \STATE $\texttt{answers[ID]} \leftarrow \mathcal{Q}(s,\texttt{question})$
            \ENDIF
        \ENDFOR

        \IF{$|\texttt{questions}| \geq 5$} 
            \FOR{$(\texttt{ID}, \texttt{question}) \in \texttt{questions}$}
                \IF{$\texttt{ID} \notin h$}
                    \IF{$\neg \mathcal{R}(M_\mathcal{R},p,q,h,\texttt{question})$}
                        \STATE \texttt{del questions[ID]})
                        \STATE \texttt{del answers[ID]})
                    \ENDIF
                \ENDIF
            \ENDFOR
        \ENDIF
            
        \FOR{$n\in 1...\texttt{sampleSize}$}
            \STATE $\texttt{logic} \leftarrow\mathcal{L}(M_{\mathcal{L}},p,q,\texttt{questions})$
            % \STATE $\texttt{answer} \leftarrow \text{evaluate\_formula}(\texttt{logic}, \texttt{valuation})$
            \IF{$\texttt{answer}$ is maybe}
                % \STATE Parse $\texttt{logic}$ to get it's parse tree and sub-formulas.
                \STATE Prune the parse tree such that it only contains sub-formulas evaluating to maybe. 
                \STATE BFS through the pruned parse tree and assign to \texttt{answer} the basic question corresponding to the first ID encountered in the parse tree
            \ENDIF
            \STATE Add $(\texttt{logic}, \texttt{answer})$ to \texttt{samples}
        \ENDFOR
        \STATE $\texttt{answer} \leftarrow \textbf{self\_consistency}(\texttt{samples})$
        \RETURN \texttt{answer}
    \end{algorithmic}\label{alg:pipeline}
\end{algorithm}

\section{Experiments and Analysis}\label{sec:evaluation}

This section details the experiments and analysis used to evaluate the proposed approach.

\subsection{Datasets}
As described in Section \ref{sec:related-work:cmr}, the ShARC and QA4PC datasets enable the assessment of conversational machine reading and policy compliance detection systems. We take advantage of the curated expression trees and policy questions in QA4PC to decide which model to use in our pipeline and we use ShARC to assess the overall effectiveness of our pipeline. Importantly, our question decomposition and logic formulation modules are not trained on these datasets, but perform inference from in-context learning.

\subsection{Model Choice}
Models were selected for the pipeline based on their ability to perform the policy decomposition and logic formulation tasks. We chose to test instruct and code models that are publicly available and have permissive licenses. The models we tested were the 8b, and 70b llama3 instruct models \citep{IntroducingMetaLlama}, the 34b codellama model  \citep{roziereCodeLlamaOpen2024}, and the 8x7B mixtral instruct model \cite{jiangMixtralExperts2024} (see Appendix for model URLs). The rationale for including code models is that the logic formulation step requires manipulation of abstract symbols, where code models may excel. The models were evaluated against the 193 QA4PC hand-annotated training examples \cite{saeidi2021cross}.

In the first experiment, the models were given the policy and the questions needed to ask the user. The models were tasked with organizing the questions into a logical formula that best represented the policy. The models were measured by the proportion of the 193 annotated examples for which they generated a propositional logical formula that is equivalent to the ground truth formula.

In the second experiment, the models were only given the policy and asked to perform both the question decomposition and logical formulation tasks. Again, the model output was evaluated using logical equivalence. For both experiments, we also tested how the number of in-context examples affects performance. Results were collected over 5 separate runs. In a run, for each $k \in \{0,1,3,5,9,15,20\}$ the prompt for the LLM is constructed from $k$ in-context examples, which are sampled randomly from a pool of 20 handcrafted examples that do not appear in the test set.

% The evaluation is performed against 193 annotated ShARC train items provided by \cite{kotonya2022policy}. Each item is annotated with the labeled gold questions (i.e. the basic questions that are needed answer the user question) and the propositional logic formula that combines the question label into an expression that can be evaluated to determine the answer to the main question.
% Each line shows the mean and standard deviation over 5 separate runs. 

\begin{figure}[h]
\centering
\includegraphics[width=\linewidth, clip]{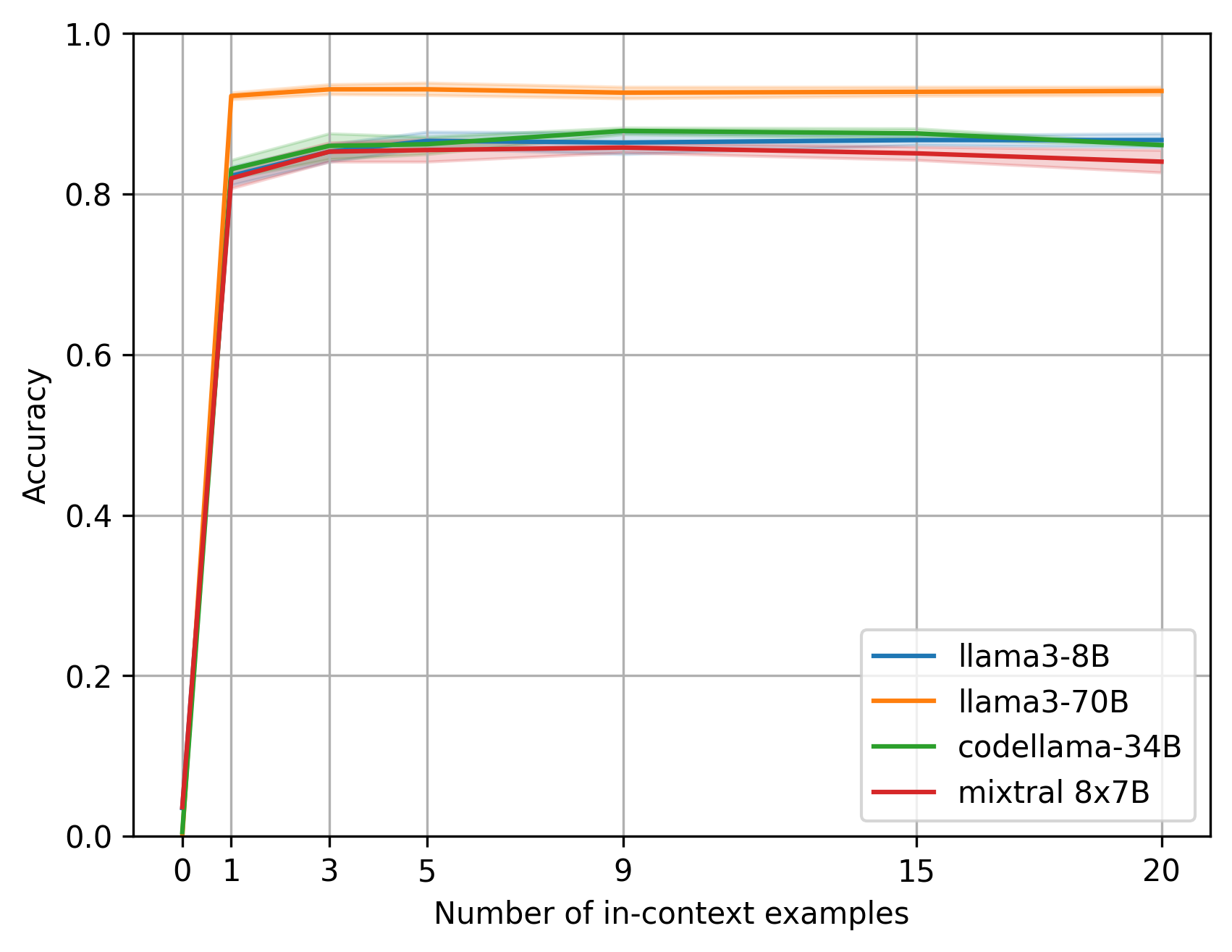}
\caption{Logic formulation given questions}
\label{fig:lf_ggq}
\end{figure}

Figure \ref{fig:lf_ggq} shows that llama3-70B performed the best for the first experiment, with an accuracy score of approximately 92\%. The llama3-8B achieved an accuracy score of approximately 85\%. This is notable because llama3-8B is significantly smaller than llama3-70B. Additionally, llama3-8B performed similarly to codellama (a 34-billion parameter model) and mixtral (which uses 13B active parameters during inference). %The 20B granite model was the worst performing model by a wide margin, approximately 20\%. The graph also shows that the models were not positively affected by increasing the number of in-context examples, expect for granite 20B.
The graph also shows that the models were not positively affected by increasing the number of in-context examples.

\begin{figure}[h]
\centering
\includegraphics[width=\linewidth, clip]{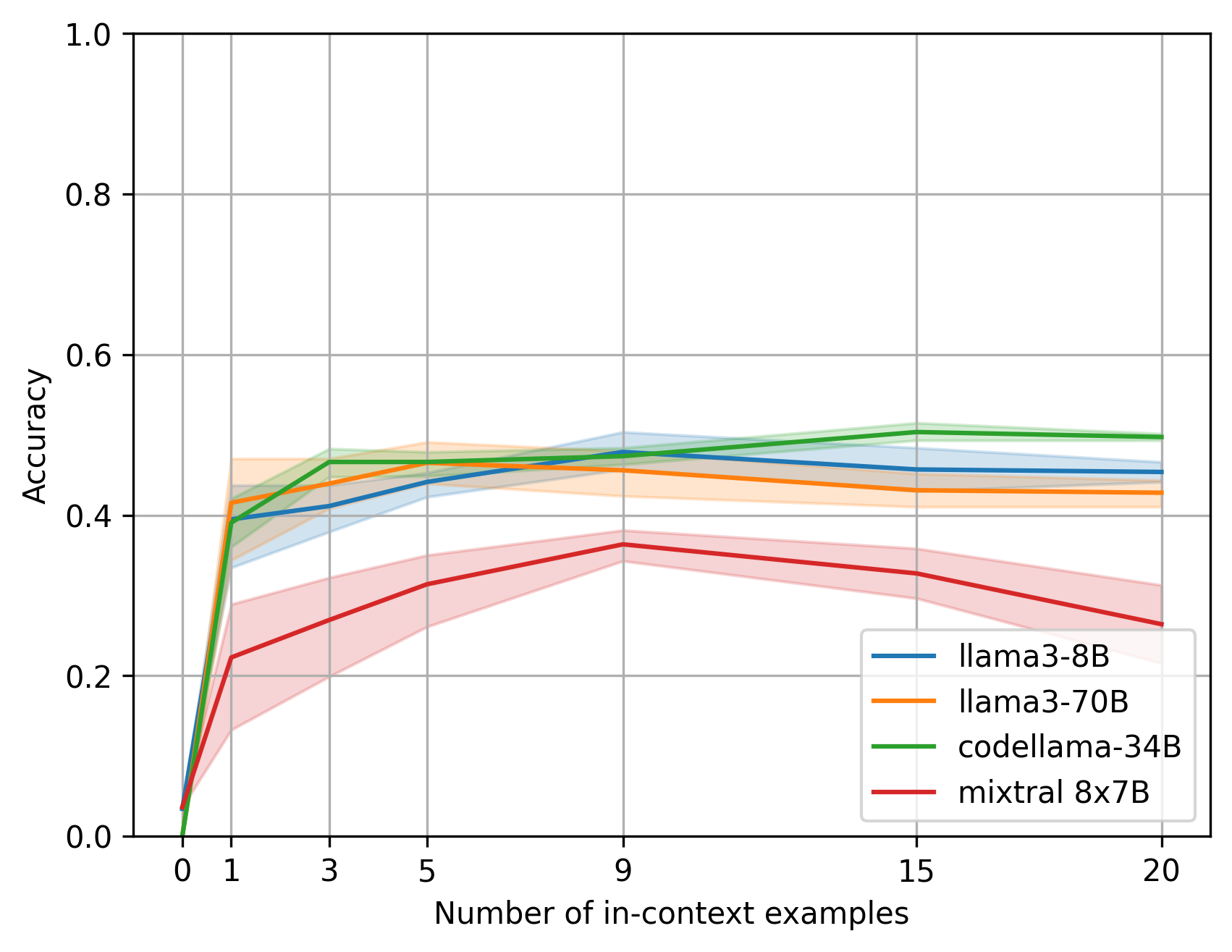}
\caption{Question decomposition and logic formulation}
\label{fig:lf_e2e}
\end{figure}

Figure \ref{fig:lf_e2e} shows the results of the second experiment, where the models were tasked with performing question decomposition and logical formulation. The accuracy of the models decreased, as expected since it is a more complex task for the models to perform. Llama3-8B, llama3-70B and codellama performed similarly around 40\% to 45\%, with codellama performing slightly better as the number of prompt examples increase. Mixtral %and granite 
performed worse than the other models and performed worse as the number of in-context examples increased. 

These experiments show that llama3-8B, llama3-70B, and codellama are suitable choices for the question decomposition and logic formulation tasks. The decision was made to use llama3-70B because it performed better than other models for the logic formulation task, and inference was faster than codellama.

\subsection{ShARC Results}

Table \ref{tab:results:accuracy} presents micro and macro accuracy scores for the ShARC dev and test sets. Our approach achieves near state-of-the-art performance, with a micro accuracy score of 79.0\%, just 1.5\% behind BiAE's 80.5\%. For macro accuracy, our approach ranks fourth with a score of 79.7\%, 3.5\% behind BiAE. Our approach showed a decline in performance on the test set compared to the development set. Specifically, the micro accuracy dropped from 79.0 to 70.2, and the macro accuracy dropped from 79.9 to 72.8. Despite this decline, our approach still performed well, ranking fourth in micro accuracy and fifth in macro accuracy among all the models tested. Furthermore, our approach uses 20 in-context examples where as the approaches that we compared against have been trained on thousands of examples.

\begin{table}[]
	\centering
    \small
	\begin{tabular}{lrrrr}
		\toprule
		\multicolumn{1}{c}{\multirow{2}{*}{\textbf{Model}}} & \multicolumn{2}{c}{\textbf{Dev set}} & \multicolumn{2}{c}{\textbf{Test set}} \\
        \multicolumn{1}{c}{}                                 & \textbf{Micro}   & \textbf{Macro}  & \textbf{Micro}   & \textbf{Macro}    \\ \midrule
		NMT             & -              & -              & 44.8           & 42.8           \\
		CM              & -              & -              & 61.9           & 68.9           \\
		BERTQA          & 68.6           & 73.7           & 63.6           & 70.8           \\
		UcraNet         & -              & -              & 65.1           & 71.2           \\
		BiSon           & 66.0           & 70.8           & 66.9           & 71.6           \\
		$\text{E}^3$    & 68.0           & 73.4           & 67.7           & 73.3           \\
		EMT             & 73.2           & 78.3           & 69.1           & 74.6           \\
		DISCERN         & 74.9           & 79.8           & 73.2           & 78.3           \\
		DGM             & 78.6           & 82.2           & 77.4           & \textbf{81.2}           \\
		BiAE            & \textbf{80.5}  & \textbf{83.2}  & \textbf{77.9}           & 81.1           \\ \midrule
		Ours            & 79.0           & 79.7           & 70.2          &   72.8            \\ \bottomrule
	\end{tabular}
	\caption{Micro accuracy and macro accuracy scores on the dev set and test set of the ShARC end-to-end task.}
	\label{tab:results:accuracy}
\end{table}

Table \ref{tab:results:bleu} presents the BLEU scores for our approach alongside those of other methods. BLEU scores measure the positional similarity of the predicted follow-up questions with the expected follow-up questions. Unfortunately, our approach performed poorly in terms of BLEU scores. Specifically, it achieved the third lowest score for BLEU-1 and the lowest score for BLEU-4 on the dev set, and the second lowest score for both BLEU-1 and BLEU-4 on the test set. However, it's important to note that our approach uses prompting and was not fine-tuned to the data like the other methods. As a result, it could not fully capture the style of follow-up questions in the data. Additionally, BLEU's sensitivity to word order leads to issues when comparing two follow-up questions. This sensitivity overemphasize word-level similarity, and neglects other important aspects of text similarity, such as semantic meaning or context.

\begin{table}[]
    \small
	\centering
	\begin{tabular}{lrrrr}
		\toprule
		\multicolumn{1}{c}{\multirow{2}{*}{\textbf{Model}}} & \multicolumn{2}{c}{\textbf{Dev set}} & \multicolumn{2}{c}{\textbf{Test set}} \\
        \multicolumn{1}{c}{}                                 & \textbf{BLEU-1}   & \textbf{BLEU-4}  & \textbf{BLEU-1}   & \textbf{BLEU-4}    \\ \midrule
		NMT             & -              & -              & 34.0           & 7.8            \\
		CM              & -              & -              & 54.4           & 34.4           \\
		BERTQA          & 47.4           & 54.0           & 46.2           & 36.3           \\
		UcraNet         & -              & -              & 60.5           & 46.1           \\
		BiSon           & 46.6           & 54.1           & 58.8           & 44.3           \\
		$\text{E}^3$    & 67.1           & 53.7           & 54.1           & 38.7           \\
		EMT             & 67.5           & 53.2           & 63.9           & 49.9          \\
		DISCERN         & 65.7           & 52.4           & 64.0           & 49.1           \\ 
		DGM             & \textbf{71.8}  & \textbf{60.2}  & 63.3           & 48.4           \\
		BiAE            & -              & -              &  \textbf{64.7}  & \textbf{51.6}          \\ \midrule
		Ours            & 57.0           & 43.9           & 46.9  & 23.3 \\ \bottomrule
	\end{tabular}
	\caption{BLEU scores on the dev set and test set of the ShARC end-to-end task.}
	\label{tab:results:bleu}
\end{table}

\subsection{Error Analysis}

To better understand the limitations of our approach, we conducted an error analysis on a randomly selected subset of items in the ShARC dev set for which our model made incorrect class predictions. For each of these items, we looked for point(s) of failure by manually inspecting the outputs of each of the following steps in our system: question-policy relevance, policy decomposition, question answering, and logic formulation. We did this for 160 randomly selected errors out of 470. 

The points of failure from most common to least common were: logic formulation (29\%), QA (28\%), question decomposition (19\%), question-policy relevance (i.e. incorrectly classifying a relevant question as irrelevant or vice versa, 5\%).\footnote{Note that this is not inconsistent with the best accuracy metrics in Figure \ref{fig:lf_ggq} because the rate of logic formulation errors from this analysis comes from a different distribution than the data used in our Model Choice experiments.} %This may appear surprising considering that we used accuracy on QA4PC expression trees to choose which language model performed the logic formulation step most accurately, and that the accuracy of most models we tested was over 80\% (Figure \ref{fig:lf_ggq}). However, it is important to note that the rate of logic formulation errors from this analysis comes from a different distribution than the data used in our Model Choice experiments. In Figure \ref{fig:lf_ggq} we report accuracy on the subset of QA4PC items with manually curated expression trees, whereas this error analysis only includes a subset of ShARC dev set items that our model evaluated incorrectly. If our error analysis subset is representative of all dev set errors, then a good estimate of logical formulation errors in the entire dev set is approximately the logical formulation error rate in our subset multiplied by the micro error rate of the entire dev set. That figure is 6\% of the entire dev set, which is not inconsistent with the best accuracy metrics in Figure \ref{fig:lf_ggq}.

Across the dev set as a whole, we found that the logic formulation results across multiple self-consistency runs were more homogeneous when the model made a correct prediction. Table \ref{tab:lf_dist} illustrates the diversity of logic formulation responses when using a self-consistency sample size of 3: unanimous (all logical formulae were logically equivalent), majority (2 out of 3 were logically equivalent), and split (each formula was logically distinct). Although there is a significant difference between the diversity in correct and incorrect predictions ($T=9.05, p<0.001, df=2144.0$), the incidence of unanimous/homogenous self-consistency samples was generally high (97.5\% and 87.4\% for correct and incorrect predictions respectively).  

\begin{table}[]
	\centering
        \small
	\begin{tabular}{lrr}
            \toprule
		\textbf{Diversity} & \textbf{Correct} & \textbf{Incorrect} \\
            \midrule
		Unanimous & 1,644 (97.5\%)    & 402 (87.4\%)    \\
		  Majority    & 40 (2.4\%) & 56 (12.2\%)           \\
		Split & 2 (0.1\%) & 2 (0.4\%)           \\
            \bottomrule
		\end{tabular}
	\caption{Distribution of self-consistency samples for logic formulation.}
	\label{tab:lf_dist}
\end{table}

Surprisingly, we found that 36\% of the items in our error analysis were borderline cases that, although different from the expected answer in the ShARC dataset, could be considered correct. %This points to the often subjective nature of these questions and conversational QA in general. Although mismatches between our model predictions and the expected answers could be attributed to different parts of our pipeline, 
We observed that a sizeable proportion of errors were due to the ShARC reference answers relying on questionable assumptions. There were also cases where the policy snippet was missing or vague, though this was far less common (3\% of inspected errors). 

For instance, in the example shown in Figure \ref{fig:overview}, our system responds with a follow up question because the policy clearly states that SBA is available to repair or replace a primary residence or personal property. However, the reference answer assumes that both of these conditions are fulfilled, despite their being no explicit evidence for it in the user scenario or in the chat history. Whether or not this assumption is justified may be subjective. Similarly, several items refer to a policy that outlines tax exemption rules for items sold to charity. In many instances, the reference answer assumed that all items mentioned by the user were being sold to charity (without evidence), while our system would respond with the question ``Are you selling the item to an eligible charity?'' In fact, across all dev set items, each time our system predicted this response, it was scored as incorrect with respect to the reference answer. These ambiguities are unavoidable to some extent, and relate to whether or not a system uses a closed world assumption in its reasoning.

\section{Conclusion}\label{sec:conclusion}

In this paper we introduce Logical Decomposition for Policy Compliance (LDPC). We first test the ability of multiple language models on the policy decomposition and logical formulation tasks using the QA4PC dataset. We then use the best identified model to build a pipeline that uses few-shot in-context learning to decompose policies into yes/no questions and assemble them into logical formulae for answering policy-related questions in a conversational setting. We additionally use a self-consistency decoding technique for producing logical formulae. We achieve competitive results on the ShARC dataset. However, we also observe that ShARC contains ambiguous items with questionable reference answers, underscoring the subjective nature of these questions and the need for careful scrutiny of LLM reasoning benchmarks.

% However, we also observe there ambiguous items with questionable reference answers are common in ShARC, highlighting the often subjective nature of these questions and the importance of carefully inspecting the results of benchmarks that claim to assess reasoning in LLMs. 

\clearpage
\section{Limitations}
Our system demonstrates that in-context learning is sufficient for achieving competitive performance on ShARC, but limitations of this approach include the simplicity of some of our system components and limitations of the ShARC dataset itself.

\paragraph{Self-consistency} The purpose of this decoding strategy is to increase diversity of reasoning paths under the hypothesis that correct answers can be reached in a variety of ways. However, as shown in our error analysis, the vast majority of logic formulations for each dev set item were homogeneous. %Although this was less frequent for items that the system got incorrect, the rate of totally uniform logic formulations for incorrect items was still almost 90\%. This does not support any claim that self-consistency made a big impact on our approach. 
This question could be further explored with other kinds of non-greedy decoding. % and greater number of self-consistency samples. 

\paragraph{Limited expressivity of three-valued logic} Although we believe the simplicity of three-valued logic to be a feature in an initial implementation, it is likely that more complex policies and scenarios require greater expressivity.
% We already saw glimpses of this ShARC. For example, when policies described hypothetical scenarios, resulting in nonsensical questions being included in the logical formula, such as ``Did you die after you've reached the State Pension age?'' Similarly, the system had difficulty expressing the intent to fulfill a condition in the future (as opposed to the condition being fulfilled at the time of asking the question). To better understand these issues, our approach (or similar) would need to be applied to more challenging and complex datasets.

\paragraph{Simple QA Model} We opted for a very common approach of using a language model fine-tuned on natural language inference. However, question answering errors were common in our analysis, occurring almost as frequently as logic formulation errors. This suggests that there is still room for improvement in this area. More powerful QA models can be explored.

%Importantly, this hinges on the complexity of the questions generated in the policy decomposition stage. 

\paragraph{Challenges to deployment} Although our system's competitive performance on ShARC without ShARC-specific fine-tuning for (de)composition is noteworthy, the results do not suggest this approach for readily use in a real-world application. % We do not believe it is possible to safely
Deploying such an application could have a serious impact on people's lives, such as determining eligibility for services or benefits. %, without
Possible mitigation methods include labor intensive fine-tuning, alignment, and implementation of other task- domain- or culturally-specific safeguards.

\paragraph{Data ambiguity issues} Lastly, our error analysis findings on the frequency of borderline instances in ShARC are consistent with prior work on spurious patterns in the dataset \cite{vermaneural}. Our approach is partly shielded from those spurious patterns owing to the lack of task-specific fine-tuning. However, this certainly limits the extent to which we can make general claims based on ShARC-based metrics.

% \clearpage
\bibliography{ref,custom}

\appendix

\label{sec:appendix}
\section{Prompt Formats}
At the policy decomposition (Section \ref{sec:proposed:pd}) and logic formulation (Section \ref{sec:proposed:lf}) steps of the pipeline, we provide each model with prompts that follow the following template: 

\indent\texttt{[INSTRUCTION]\newline\indent[IN-CONTEXT EXAMPLES]\newline\indent[PARTIAL EXAMPLE]}

\noindent 
For policy decomposition, the instruction was:

\textit{"You are given a policy and an input question. You must think logically. Do not extrapolate. Your job is to decompose the policy into its basic rules. Rephrase the rules in the form of basic questions that need to be answered to answer the input question. Here are some examples to help guide you to do what I want."}

For logic formulation, the instruction was:

\textit{"You are a highly rational and concise assistant. You are given a policy, input question and a decomposition of the policy into the basic rules in the form of questions that need to be answered to answer the input question. Please combine the question variables into a python boolean expression that can be evaluated to answer the input question."}

Table \ref{Tbl:TreeIDs} provides a list of tree IDs from the ShARC train set that were used as in-context examples. Additionally, examples 4 and 7 were manually crafted to illustrate policies with complex logical.

\begin{table}[h]
\centering
\small
\caption{In-context examples used}
\label{Tbl:TreeIDs}
\begin{tabular}{ll}
\toprule
ID & Tree ID \\ \midrule
1 & d752ee0af2d200afb785cbe9ee99f0ffc04c28a0 \\
2 & 7060f13b78f1c604088f35f8edb3a6f88b4ae45a \\
3 & 55fe163fab637ebe7dfc2557ce08d1bc7669d106 \\
4 &  Handcrafted \\
5 & f63a9029b95e264424fce797c4e627701913209b \\
6 & 45c0b83ec88820c3c9c4a8d3b94d1051fce9e3ec \\
7 &  Handcrafted \\
8 & 54bbdc5cdea357390ec8d11c13d1c4c26238074b \\
9 & d5015eb79f2ab92950bb9c84cf5eddaf45cd47eb \\
10 & 15b63a500f40395913452861581a0a14b8e6ca25 \\
11 & 21fe1028028910dd3f9b4071a3135639efedc236 \\
12 & b7fa1f12f7ec1b6cff4267e2f4b2a8b8dbaa05fe \\
13 & 21f3e298787fd4984709a26a1b7b5201ef5b4e59 \\
14 & 3e276836bcd6e90e63105d41a14d6b65b21b487d \\
15 & 58b139a8f03204de8ab53e9b407f16f99a60a326 \\
16 & 2b48a053981dc4c15c730751c5fbe21ba75b3aa9 \\
17 & 3e6b7f945d0e8a8c090987926a819aaa40d44ed0 \\
18 & e579aa599ceda30399379976383f28979760dc8e \\
19 & b6c4e73a9754b24c36de6c476ec0abc1385f857a \\
20 & a50904bdf5f21b9d625280122573cc8e1e3f090b \\ \bottomrule
\end{tabular}
\end{table}

\subsection{Publicly Available Models}

\begin{itemize}
    \item \url{https://huggingface.co/meta-llama/Meta-Llama-3-8B-Instruct}
    \item \url{https://huggingface.co/meta-llama/Meta-Llama-3-70B-Instruct}
    \item \url{https://huggingface.co/mistralai/Mixtral-8x7B-Instruct-v0.1}
\end{itemize}

\end{document}